\documentclass[letterpaper, 10 pt, conference]{ieeeconf}   

\IEEEoverridecommandlockouts                              

\usepackage{color}

\usepackage{amssymb}
\usepackage{graphicx}
\usepackage{subcaption}
\usepackage{siunitx}
\usepackage{booktabs} 
\usepackage{amsmath}

\usepackage{amsthm}
\usepackage{bm}
\usepackage{multirow}
\let\labelindent\relax
\usepackage{enumitem}
\usepackage{multicol}
\usepackage[bookmarks=true]{hyperref}

\title{\LARGE \bf {Learning Continuous Control with Geometric Regularity\\from Robot Intrinsic Symmetry}}

\author{Shengchao Yan$^1$, Baohe Zhang$^1$, Yuan Zhang$^1$, Joschka Boedecker$^1$, Wolfram Burgard$^2$
\thanks{
The authors are with the $^1$Department of Computer Science, University of Freiburg, Germany, and the $^2$Department of Engineering, University of Technology Nuremberg, Germany.
}
}

\begin{document}

\maketitle
\thispagestyle{empty}
\pagestyle{empty}

\begin{abstract}
Geometric regularity, which leverages data symmetry, has been successfully incorporated into deep learning architectures such as CNNs, RNNs, GNNs, and Transformers. While this concept has been widely applied in robotics to address the curse of dimensionality when learning from high-dimensional data, the inherent reflectional and rotational symmetry of robot structures has not been adequately explored. Drawing inspiration from cooperative multi-agent reinforcement learning, we introduce novel network structures for single-agent control learning that explicitly capture these symmetries. Moreover, we investigate the relationship between the geometric prior and the concept of Parameter Sharing in multi-agent reinforcement learning. Last but not the least, we implement the proposed framework in online and offline learning methods to demonstrate its ease of use. Through experiments conducted on various challenging continuous control tasks on simulators and real robots, we highlight the significant potential of the proposed geometric regularity in enhancing robot learning capabilities.
\end{abstract}


\section{Introduction}

Robots have the ability to undertake tasks that are dangerous or difficult for humans. 
With more degrees of freedom, they can perform increasingly complex tasks. For example, humanoid robots and quadrupedal robots can walk over challenging terrain, while robot arms and hands can achieve dexterous manipulation. 
However, controlling robots with a large number of degrees of freedom becomes increasingly difficult as the observation and action space grows exponentially. 
Although deep reinforcement learning has been employed to solve various robot control problems~\cite{LillicrapHPHETS15, miki2022learning,wu2023daydreamer,allshire2022transferring}, learning effective control strategies for these robots remains a challenging task.

Training neural networks on high-dimensional data is known to be challenging due to the curse of dimensionality~\cite{bronstein2021geometric}. To overcome this challenge, researchers have developed network architectures and incorporated various inductive biases that respect the structure and symmetries of the corresponding domains.
For example, convolutional neural networks (CNNs) leverage the strong geometric prior of images by incorporating translation and rotation equivariance into the design of convolutional layers. This ensures that the extracted features move along with the original image, regardless of the direction it is shifted in.
Similarly, graph neural networks (GNNs) take advantage of the geometric prior of permutation invariance in other domains to capture the relationships among objects. 
Overall, incorporating domain-specific inductive biases and symmetries can greatly improve the ability of neural networks to learn from high-dimensional data.

\begin{figure}[t]
    \centering
    \includegraphics[width=0.48\textwidth]{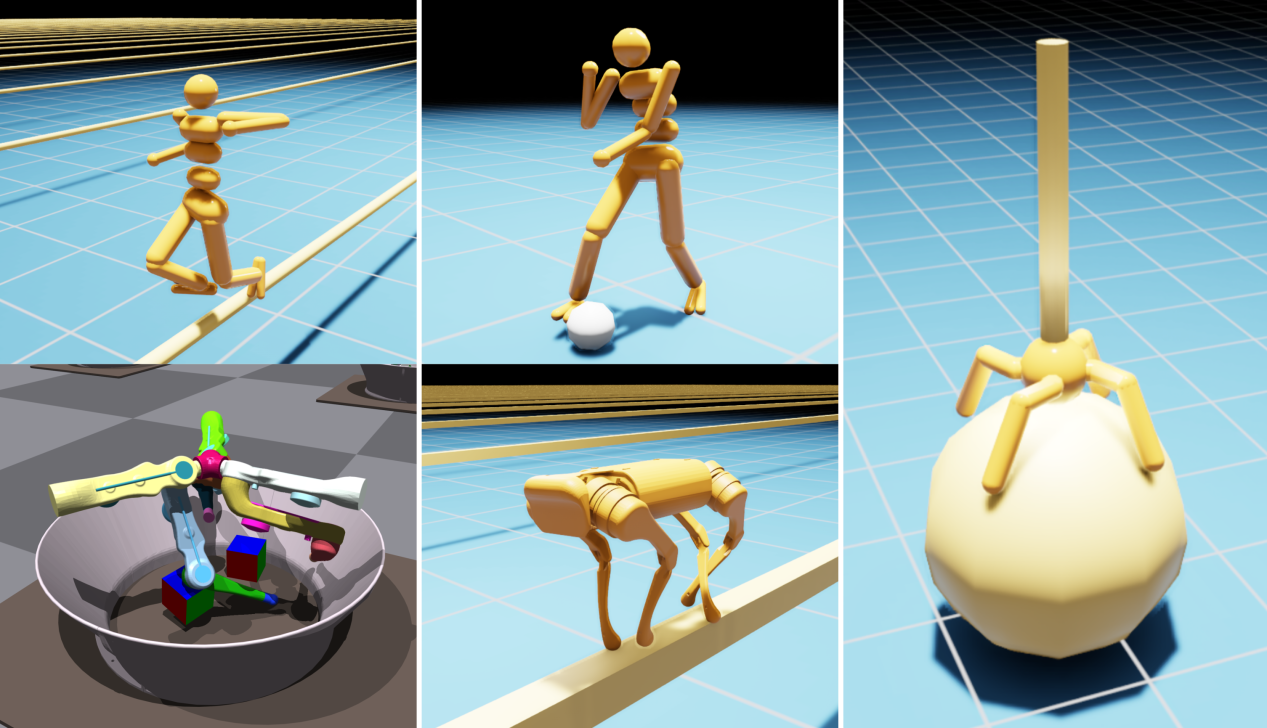}
    \caption{Tasks challenging for current deep reinforcement learning baseline algorithms.}
    \label{fig:experiments}
\end{figure}

However, in the realm of robot learning research, the potential benefits of exploiting symmetry structures present in environments, such as reflectional and rotational symmetry, remain largely unexplored. Therefore, how to combine this prior knowledge to effectively improve the existing approaches still is worth to be investigated.
To bridge the research gap, we propose to reformulate the control problems using a Multi-Agent Reinforcement Learning (MARL) framework to better leverage the symmetry structures. 

Specifically, we introduce the Multi-Agent with Symmetry Augmentation network structure (MASA), an architecture for designing control policies or value functions which leverage the transformation equivariance or invariance of the corresponding symmetric robot structure.
Instead of learning policy and critic functions in the joint action space composed of all actuators in a robot, we divide the robot into several symmetric components and learn a policy for each of them. The critic function maps observations with or without actions from all agents to a centralized value.
Additionally, we establish a connection between our proposed geometric prior and the important concept of ``Parameter Sharing'' in MARL, which drastically reduces the optimization space and speeds up the learning process. 
We demonstrate the surprising effectiveness of our approach by combining the new architecture with both online and offline model-free deep learning methods.
We evaluate the proposed method on a set of challenging robot control tasks (see Fig.~\ref{fig:experiments}). The experimental results demonstrate that our method significantly improves the performance and data efficiency of robot control learning tasks.

\section{Background and Related Work}
\label{sec:background_and_related_work}

\subsection{Multi-Agent Reinforcement Learning (MARL)}
MARL is an extended reinforcement learning method for decision-making problems, where multiple agents can interact and learn in one environment simultaneously. The common mathematical framework for MARL problems is Markov games. A Markov game is a tuple $\left\langle \mathcal{N}, \mathcal{S}, \mathcal{O}, \mathcal{A}, P, R_{i}, \gamma \right\rangle$. $\mathcal{N}$ is the set of all agents and $\mathcal{S}$ is the set of states. $\mathcal{O}_i$ and $\mathcal{A}_i$ are observation space and action space for agent $i$, while $\mathcal{O} = {\times}_{i \in \mathcal{N}} \mathcal{O}_{i}$ and $\mathcal{A} = {\times}_{i \in \mathcal{N}} \mathcal{A}_{i}$ represent joint observation space and joint action space. Define $\Delta_{\mathcal{S}}$ and $\Delta_{\mathcal{A}}$ be the probability measure on $\mathcal{S}$ and $\mathcal{A}$ respectively. Then $P$ is the transition probability $P(s'|s,a): \mathcal{S} \times \mathcal{A} \to \Delta_{\mathcal{S}}$. Each agent $i$ maintains an individual reward function $R_i(s,a): \mathcal{S} \times \mathcal{A} \to \mathbb{R}$, and the future rewards are discounted by the discount factor $\gamma \in [0,1]$. Let $\Pi_i = \left\{ \pi_i(a_i|o_i): \mathcal{O}_i \to \Delta_{\mathcal{A}_i} \right\}$ be the policy space for agent $i$, then the objective for agent $i$ is represented as $\max_{\pi_i} \mathbb{E}_{\pi, P} \left[ \sum_{t=0}^{+\infty} \gamma^t R_i(s_t, a_t) \right]$. In practice, the state space and the observation space can be identical if the observation has already fully described the system. Our paper also follows this assumption and uses observation alone. 

Multi-Agent Mujoco~\cite{peng2021facmac} is a popular benchmark for MARL algorithms which divides a single robot into several distinct parts with separate action space. However, the state-of-the-art MARL algorithms still couldn't match the performance of the single-agent algorithms on this benchmark.
Different from their work, in which they arbitrarily divide robots into parts and ignore the geometric structures of the robots, we leverage ideas from geometric regularity during the MARL training and our results show that MARL can outperform single-agent algorithms by a substantial margin.

\subsection{Symmetry in Robot Learning}
In the robot learning domain, two groups of symmetric structures have been used to improve performance and learning efficiency.
1) \textbf{Extrinsic Symmetry}:
By extrinsic symmetry we refer to the symmetries existing in the exteroceptive sensors of the robot such as camera input.
Some works~\cite{wang2022robot,zhu2022grasp,wang2022so,wang2022equivariant} have been proposed to integrate these symmetries into system identification approaches with neural networks, especially CNN-structured ones.
These methods can largely improve the performance for manipulation tasks, but they are mostly used for manipulation tasks with image input and grippers without roll-pitch movement.
Learning symmetry in the latent space  directly from data~\cite{mondal2022eqr} is still limited to representation learning from images.
2) \textbf{Intrinsic Symmetry}:
Different from extrinsic symmetries, intrinsic symmetries mostly naturally come from the physical constraints in the control system.
For example, a humanoid robot control task exhibits reflectional symmetry. A symmetric control policy on such robot is usually more natural and effective.
A data-augmentation method~\cite{mavalankar2020goal} is proposed to improve reinforcement learning methods for rotation invariant locomotion.
To directly incorporate symmetry in the policy, it is also proposed to numerically construct equivariant network layers~\cite{van2020mdp,ApraezMAM23}. 
However, additional calculation is required to design the network even if the domain specific transformation is given, leading to a relatively complex procedure. Moreover, the policy network of~\cite{van2020mdp} only considers a pole balancing task with discrete action and \cite{ApraezMAM23} has no experiments on robot control policy learning.
Researchers investigate four different methods to encourage symmetric motion of bipedal simulated robots~\cite{abdolhosseini2019learning}.
They are implemented via specific policy network, data augmentation or auxiliary loss function. Even though the robots' motions become natural-looking, they do not show a major improvement on different tasks.
The policy network method in \cite{abdolhosseini2019learning} is similar to ours. But instead of a specific network merely for locomotion tasks with reflectional symmetry, we propose a generic equivariant policy network for both reflectional and rotational symmetries, the predominant symmetry features in robotic systems and animal biology. Moreover, we approach the control task from the viewpoint of multi-agent systems. Finally, we achieve substantial performance improvements in our experiments by reducing the policy search space.

\section{Single robot Control as MARL}
\label{sec:single_robot_control_as_marl}

Instead of learning a single-agent policy to control the whole robot, which will lead to a large observation-action space that is difficult to optimize, we introduce multiple agents that are responsible for each individual component of the robot inspired by MARL. We further propose a framework driven by the presence of symmetry structures in many robots and exploit such inductive biases to facilitate the training by applying parameter sharing techniques. 

Our method consists of (1) identifying the geometric structures of different robots and dividing single robots into multiple parts accordingly; (2) reformulating the control problem under a MARL framework; (3) optimizing policies with Parameter Sharing. 

\subsection{Dividing Single Robots into Multiple Parts}
\label{sec:dividing_robot}

\begin{figure}[t]
    \vspace{5pt}
    \centering
    \begin{subfigure}[b]{4cm}
        \centering
        \includegraphics[width=3.2cm]{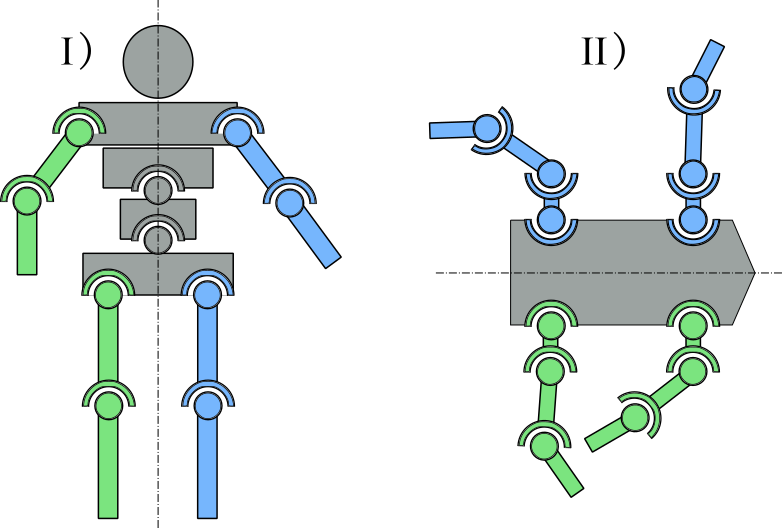}
        \caption{Reflectional symmetry}
        \label{fig:reflection}
    \end{subfigure}\hfill
    \begin{subfigure}[b]{4cm}
        \centering
        \includegraphics[width=4cm]{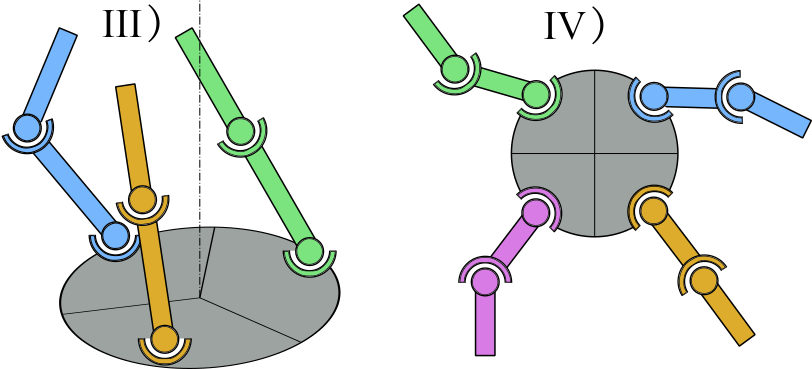}
        \caption{Rotational symmetry}
        \label{fig:rotation}
    \end{subfigure}
    \caption{\textbf{Agent partitioning considering symmetry structures}: 
    Humanoid and Cheetah robots split into left and right parts by reflectional symmetry;
    TriFinger and Ant robots split into three and four parts by rotational symmetry, where each part is controlled individually by a dedicated agent.
    The central part (grey) is controlled by all agents.}
    \label{fig:partition}
    \vspace{-5pt}
\end{figure}

\begin{figure*}[t]
    \vspace{5pt}
    \centering
    \begin{subfigure}[b]{5cm}
        \centering
        \includegraphics[width=4.8cm]{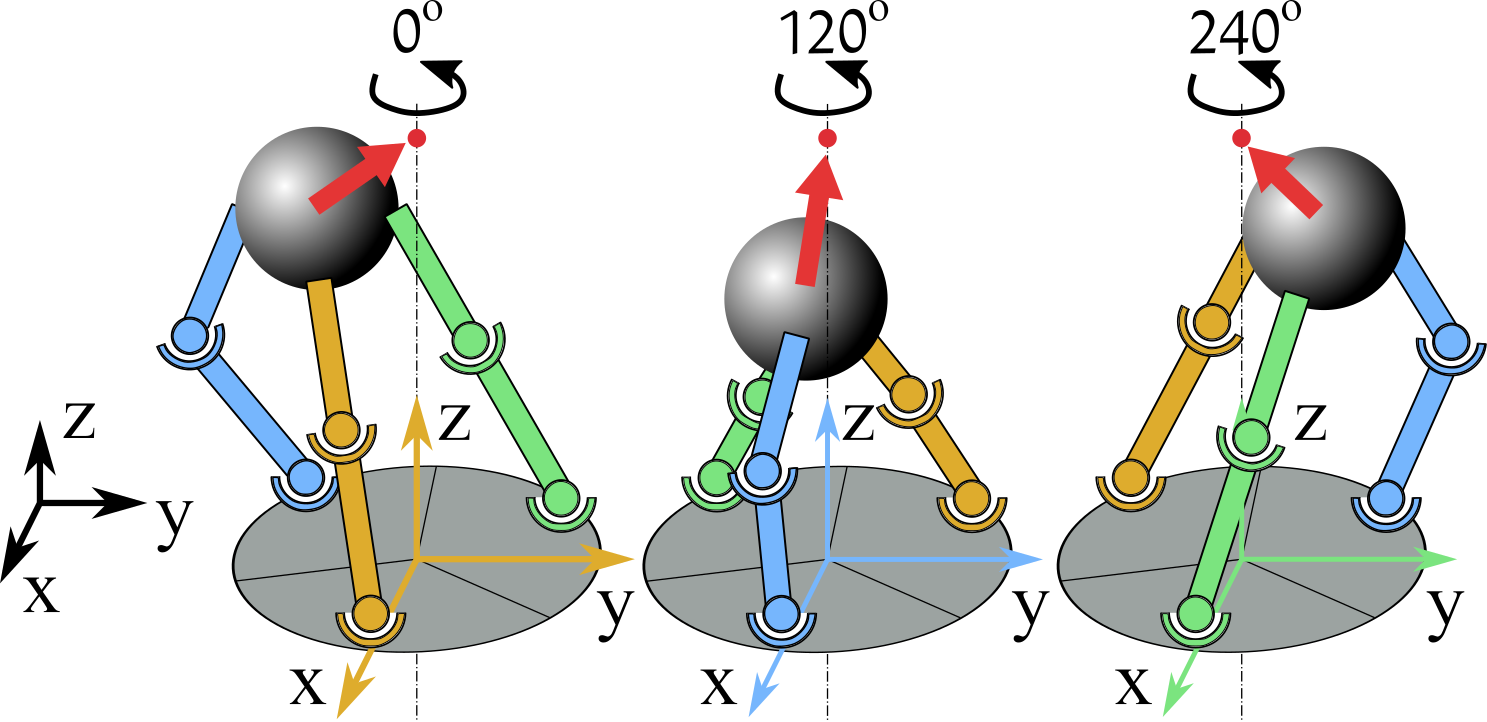}
        \caption{Symmetric states of TriFinger.}
        \label{fig:trifingerSym}
    \end{subfigure}\hfill
    \begin{subfigure}[b]{5.6cm}
        \centering
        \includegraphics[width=5.6cm]{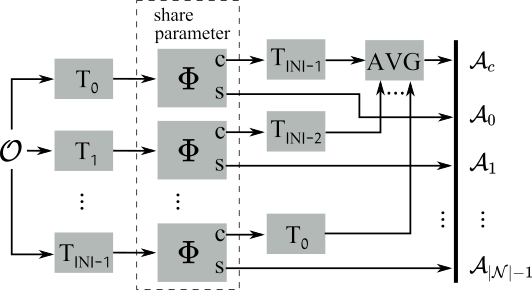}
        \caption{Policy network}
        \label{fig:policy_network}
    \end{subfigure}\hfill
    \begin{subfigure}[b]{6cm}
        \centering
        \includegraphics[width=6cm]{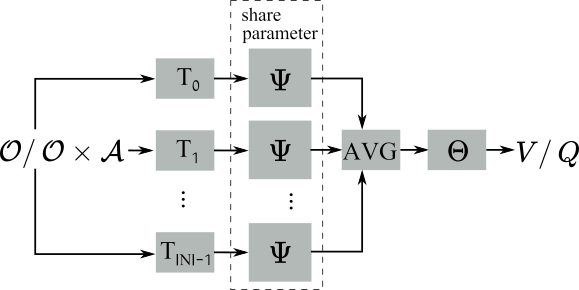}
        \caption{Critic network}
        \label{fig:critic_network}
    \end{subfigure}
    \caption{a) TriFinger robot moves an object towards a target position. The black coordinate system is the global system, while the colored ones are local systems. The red arrow represents the desired moving direction of the manipulated object. Note that the actions of different body parts should be equivariant with regard to the rotations. b) Equivariant policy network with parameter $\Phi$. \textbf{c} and \textbf{s} stand for central and symmetric actions. c) Invariant critic network with parameter $\Psi,\Theta$.}
    \label{fig:symmetricStates}
    \vspace{-5pt}
\end{figure*}

    Previous research~\cite{peng2021facmac} divides a single robot into multiple parts to evaluate the performance of MARL methods. However, its irregular partitioning makes it hard for multi-agent methods to compete with the single-agent methods. In this paper, we instead take advantage of the symmetry structures of robots.

    As shown in Fig.~\ref{fig:reflection}, robots with reflectional symmetry can be partitioned into left (blue), right (green) and a central part (grey).
    The robots with rotational symmetry in Fig.~\ref{fig:rotation} are partitioned into parts with the same number of symmetric limbs (colour) and a central part (grey).
    For a robot with any of these symmetric structures, we split the whole robot's original observation-action space $\mathcal{O}\times\mathcal{A}$ by $\mathcal{O}=\mathcal{O}_\text{c}\times\prod_{i \in \mathcal{N}} \mathcal{O}_\text{s,i}$  and $\mathcal{A}=\mathcal{A}_\text{c}\times\prod_{i \in \mathcal{N}} \mathcal{A}_\text{s,i}$. 
    $\mathcal{O}_\text{c}$ represents the central observation space, which consists of measurements that do not have symmetric counterparts, such as the position, orientation, velocity and joints of the torso, target direction, or states of the manipulated objects. Raw sensor data such as images and point clouds may also belong to central observation. 
    $\mathcal{O}_{\text{s},i}$ corresponds to symmetric observation spaces, whose measurements may include joint positions and velocities from the limbs, contact sensor measurements of the feet or fingers, and so on. The symmetric observation spaces are the same for any $i\in\mathcal{N}$ due to the robots' symmetric property. 
    $\mathcal{A}_\text{c}$ and $\mathcal{A}_{\text{s},i}$ are the action spaces for central (e.g., humanoid robot's pelvis and twist) and symmetric (e.g., limbs) robot parts.


\subsection{Multi-Agent Reinforcement Learning Formulation}
\label{sec:marl_formulation}
    Assume the original observation and action of the whole robot be $o \in \mathcal{O}$ and $a \in \mathcal{A} $ respectively and the number of agents $|\mathcal{N}|$, equal to the number of symmetry parts of the robots. For each agent $i \in \mathcal{N}$, there is a unique transformation function $\mathrm{T}_{i}\in\mathcal{T}$ to obtain its own observation $o_i = \mathrm{T}_{i}(o)$, where $\mathcal{T}$ is a set of symmetry transformation functions for observations or actions defined by the corresponding symmetric structure. We describe the transformation functions later in this section.
    Each agent generates the local action $a_i$, consisting of $a_{\text{c},i} \in \mathcal{A}_c$ and $a_{\text{s},i} \in \mathcal{A}_{s,i}$ for central and symmetric actions, by its own policy network. 
    Finally, the whole robot's action $a$ is recovered by gathering all symmetric actions $a_{s,i}$ and merging all central actions $a_{c,i}$ into $a_c$. 
    Regarding the reward function, our formulation follows the cooperative MARL setup, where $R_i$ for all $i \in \mathcal{N}$ are identical at every time step. This shared reward is calculated by a task-related reward function $R(o, a)$ which depends on the whole robot's observation and action. 
    
    We take the TriFinger robot in Fig.~\ref{fig:trifingerSym} as an example to explain the transformation set $\mathcal{T}$ for rotational symmetry.
    First, we define a local coordinate system for each of the three agents, with the origin at the center of the robot base, $z$ axis along the robot symmetry axis and $x$ axis pointing to the base joint of the corresponding limb.
    Then, we arbitrarily select an agent as the base agent. Here we take the yellow one.
    The coordinate system of the base agent is used as the global system and the robot observation $o$ should be converted into it, resulting in $\mathrm{T}_0$ as identity transformation for the base agent.
    We further describe the transformation function of other agents.
    Different observation components are categorized into different groups by $o=[o_\text{inv},o_\text{v}]$, where $o_\text{inv}$ stands for quantities invariant under the symmetry transformation, such as robot id or the delay of control systems, and $o_\text{v}$ for the variant ones.
    The basic idea of the transformation of agent $i$ is to rotate the whole environment so that its local coordinate system overlaps with the global system.
    As a result of the rotation, coordinate-system-irrelevant quantities of the three fingers shift circularly, and other variant values are changed by the rotation transformation.
    The same rules apply for the action space transformation due to the symmetric robot structure.
    Note that some observation quantities have to be both shifted and transformed, such as the fingertip position.
    Given the observation $o=[t_\text{delay},\bm{\alpha_0},\bm{\alpha_1},\bm{\alpha_2},p_\text{object}]$, the symmetric observation of agent $i$ can be calculated by $\mathrm{T}_i(o)=[t_\text{delay},\bm{\alpha_i},\bm{\alpha_{(i+1)\mod{3}}},\bm{\alpha_{(i+2)\mod{3}}},\mathrm{R}_i(p_\text{object})]$, where $t_\text{delay}$ is the control delay, $\bm{\alpha_i}$ is the joints angle position of finger $i$, $p_\text{object}$ is the object position, and $\mathrm{R}_i$ is the corresponding rotation.
    The transformation function for reflectional symmetry is defined in a similar way. The only difference is that the local coordinate systems are reflected to overlap with the global system instead of being rotated.

    We apply our method to both online and offline learning algorithms.
    To optimize the policies with interaction with the environment, we adopt the multi-agent version of Proximal Policy Optimization (PPO)~\cite{schulman2017proximal} methods. PPO is a popular model-free actor-critic reinforcement learning algorithm in different domains~\cite{miki2022learning,allshire2022transferring,yan2021courteous} for its stability, good performance and ease of implementation. Its multi-agent version also achieves competitive performance on different MARL benchmarks~\cite{yu2022surprising,de2020independent}. 
    For offline settings where the agent learns from fixed dataset without interacting with the environment, we implement the proposed framework with Behavior Cloning (BC) and Implicit Q-Learning (IQL)~\cite{kostrikov2022offline}, a state-of-the-art offline reinforcement learning algorithm.

\subsection{Geometric Regularization}
\label{sec:geometric_regularization}

\textit{Parameter Sharing} has been considered as a crucial element in MARL for efficient training~\cite{gupta2017cooperative}.
By enabling agents to share parameters in their policy networks, parameter sharing not only facilitates scalability to a large number of agents but also enables agents to leverage shared learned representations, leading to reduced training time and improved overall performance.
However, it is revealed that indiscriminately applying parameter sharing could hurt the learning process~\cite{christianos2021scaling}.
Successful parameter sharing relies on the presence of homogeneous agents as a vital requirement.
In other words, agents should execute the same action if they are given the same observation.
For our method, this is realized by the symmetry transformation.

Take the task in Fig.~\ref{fig:trifingerSym} as an example, where the manipulator with three symmetrically aligned fingers has to move the sphere towards a target position. If the whole system is rotated by $120^\circ$ or $240^\circ$ around the $z$ axis of the robot base, under the optimal policy, the actions should also shift circularly among the three fingers.
Given the whole robot's observation $o$, this relationship can be denoted by:
\begin{equation}
    A_{\text{s},j}(\mathrm{T}_i(o))=A_{\text{s},i}(\mathrm{T}_j(o)),\quad A_{\text{c}}(\mathrm{T}_i(o))=\mathrm{T}_i(A_\text{c}(o))
    \label{eq:equivariance}
\end{equation}
where $A_{\text{s},j}$ is the symmetric action of the $j$-th agent, $A_{\text{c}}$ is the central action, which exists for Humanoid robot, $\mathrm{T}_i$ is the symmetry transformation of agent $i$.
Note that the corresponding robot parts of agents can be defined arbitrarily. It does not influence the equivariance/invariance. 
Regarding the value function $V: \mathcal{O} \to \mathbb{R}$ in RL algorithms, it possesses the invariance: $V(\mathrm{T}_i(o))=V(\mathrm{T}_j(o)),\forall i,j \in \mathcal{N}$, which also holds for the Q function $Q: \mathcal{O}\times\mathcal{A} \to \mathbb{R}$.

\begin{figure*}[t]
\vspace{5pt}
\centering
\begin{subfigure}[b]{3.5cm}
    \centering
    \includegraphics[width=3.5cm]{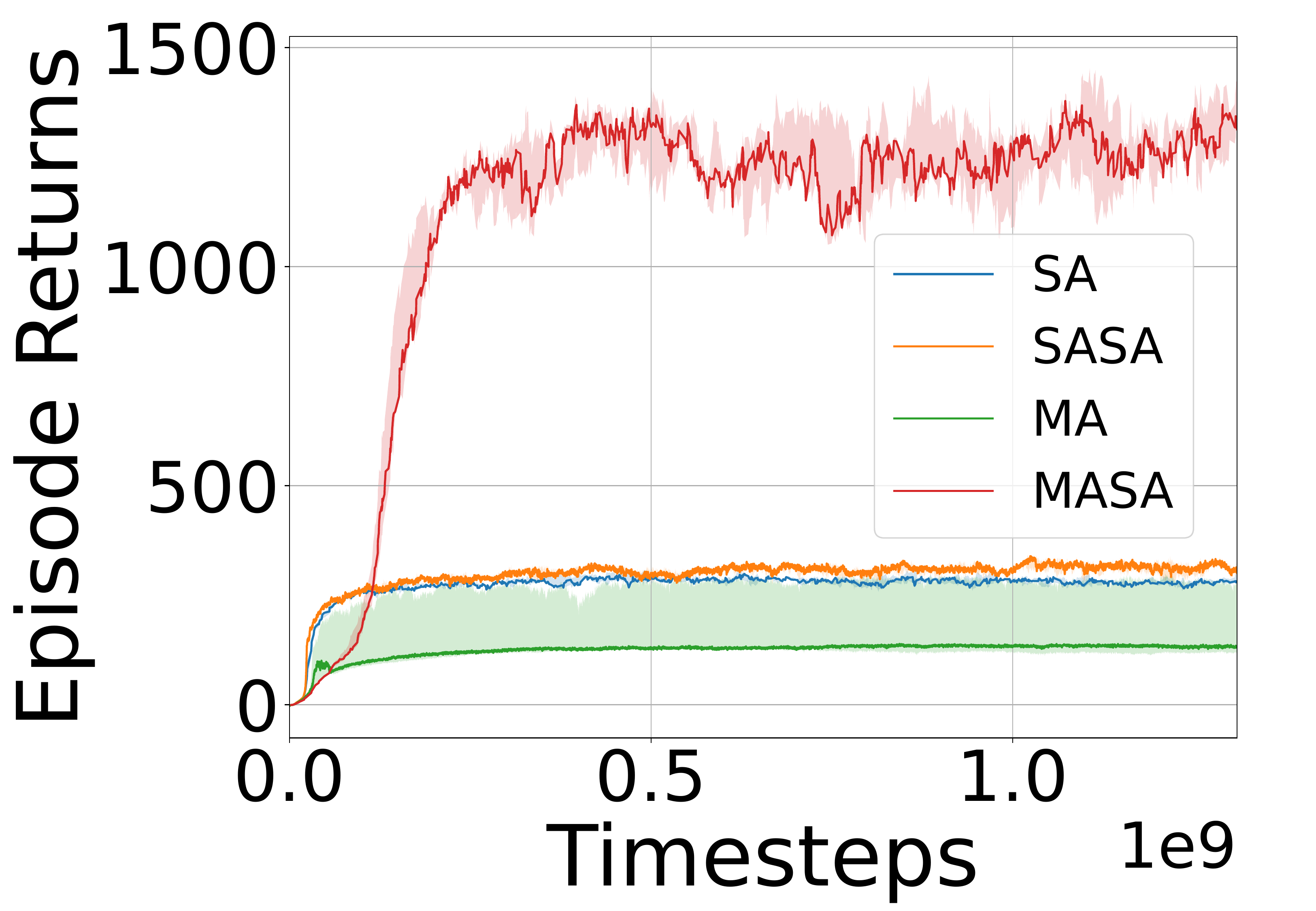}
    \caption{Humanoid Dribbling}
    \label{fig:exp_humanoid_football}
\end{subfigure}\hfill
\begin{subfigure}[b]{3.5cm}
    \centering
    \includegraphics[width=3.5cm]{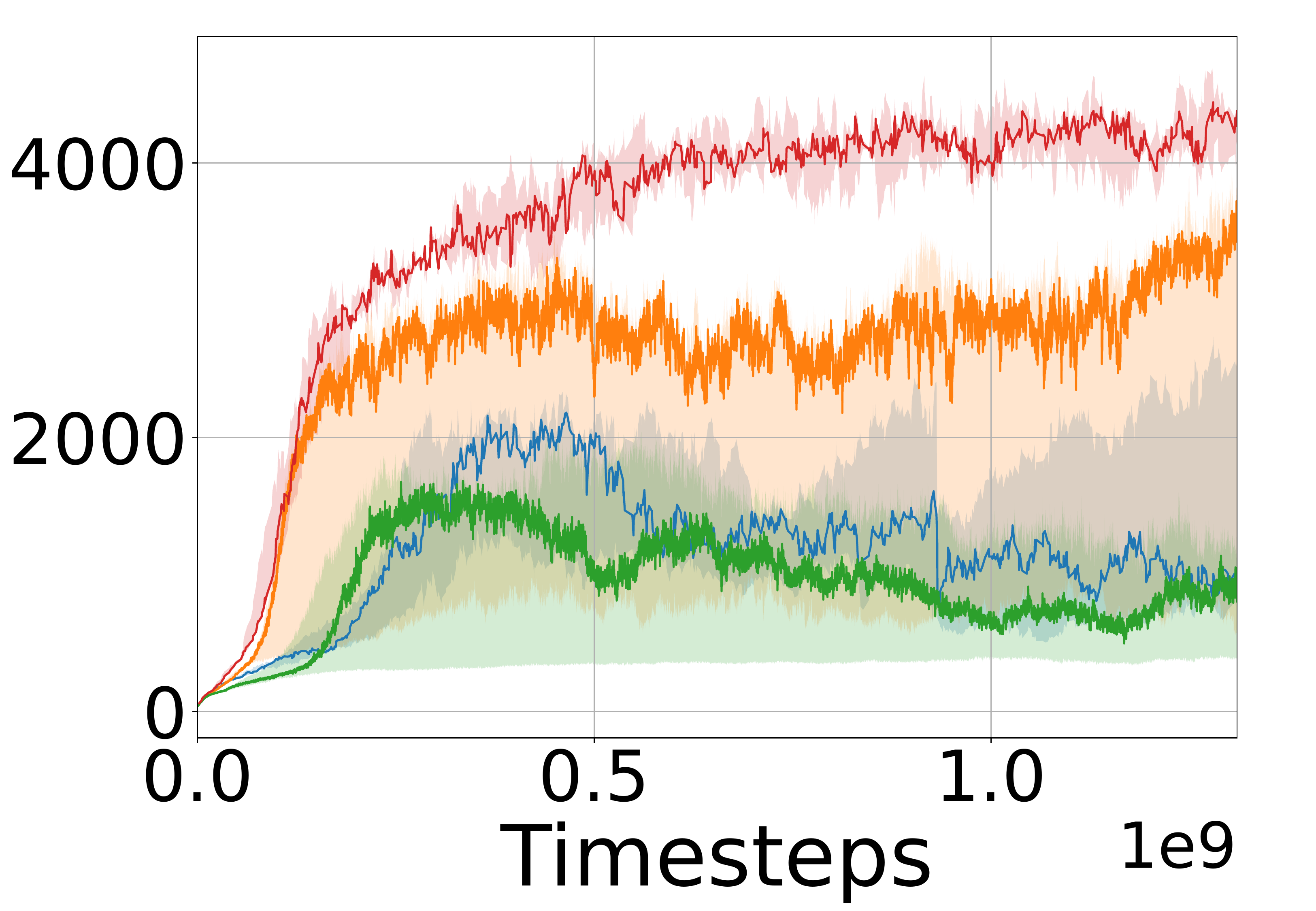}
    \caption{Humanoid Tightrope}
    \label{fig:exp_humanoid_tightrope}
\end{subfigure}\hfill
\begin{subfigure}[b]{3.5cm}
    \centering
    \includegraphics[width=3.5cm]{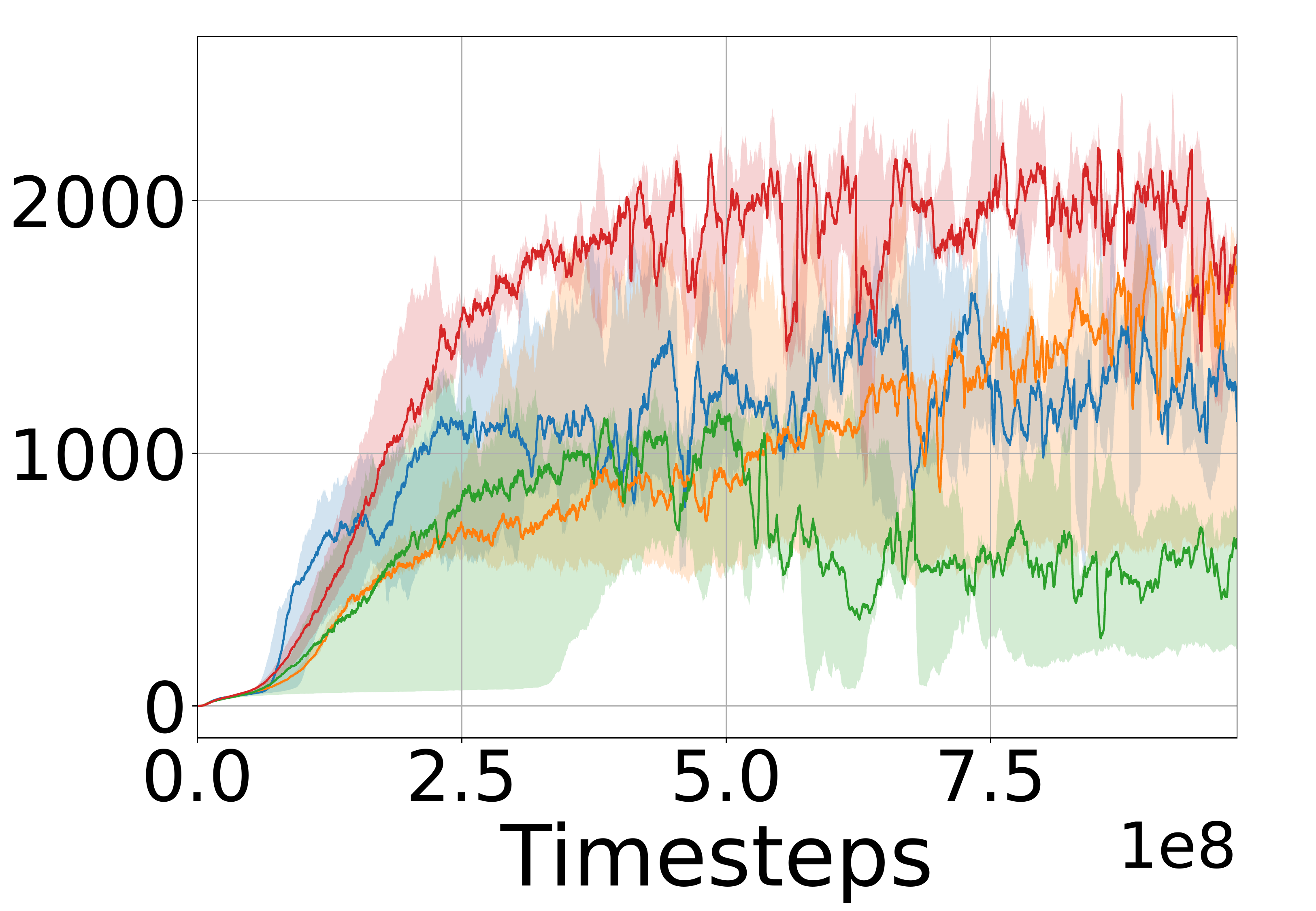}
    \caption{A1 Beam}
    \label{fig:exp_a1_beam}
\end{subfigure}\hfill
\begin{subfigure}[b]{3.5cm}
    \centering
    \includegraphics[width=3.5cm]{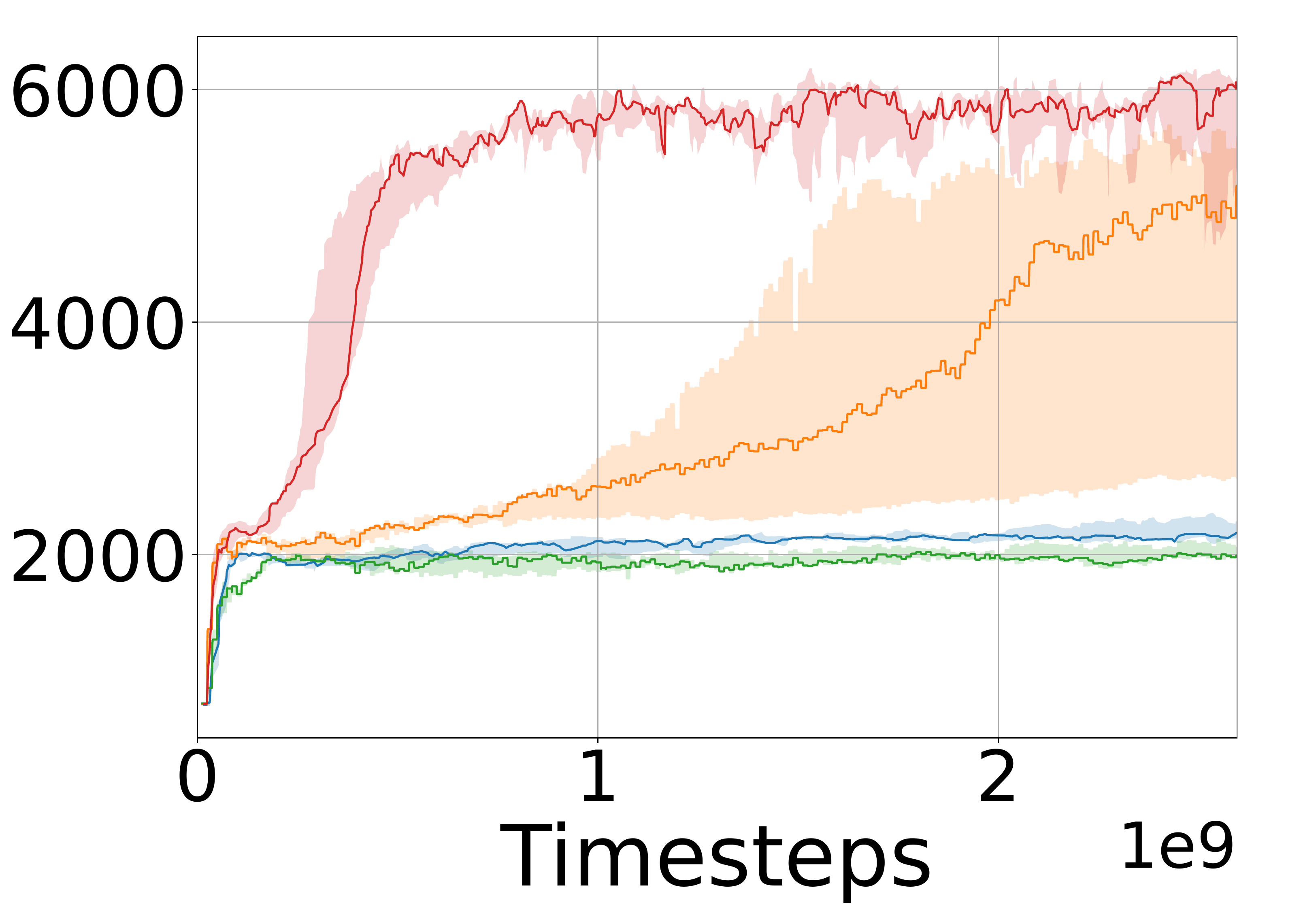}
    \caption{TriFinger Lift}
    \label{fig:exp_trifinger_move}
\end{subfigure}\hfill
\begin{subfigure}[b]{3.5cm}
    \centering
    \includegraphics[width=3.5cm]{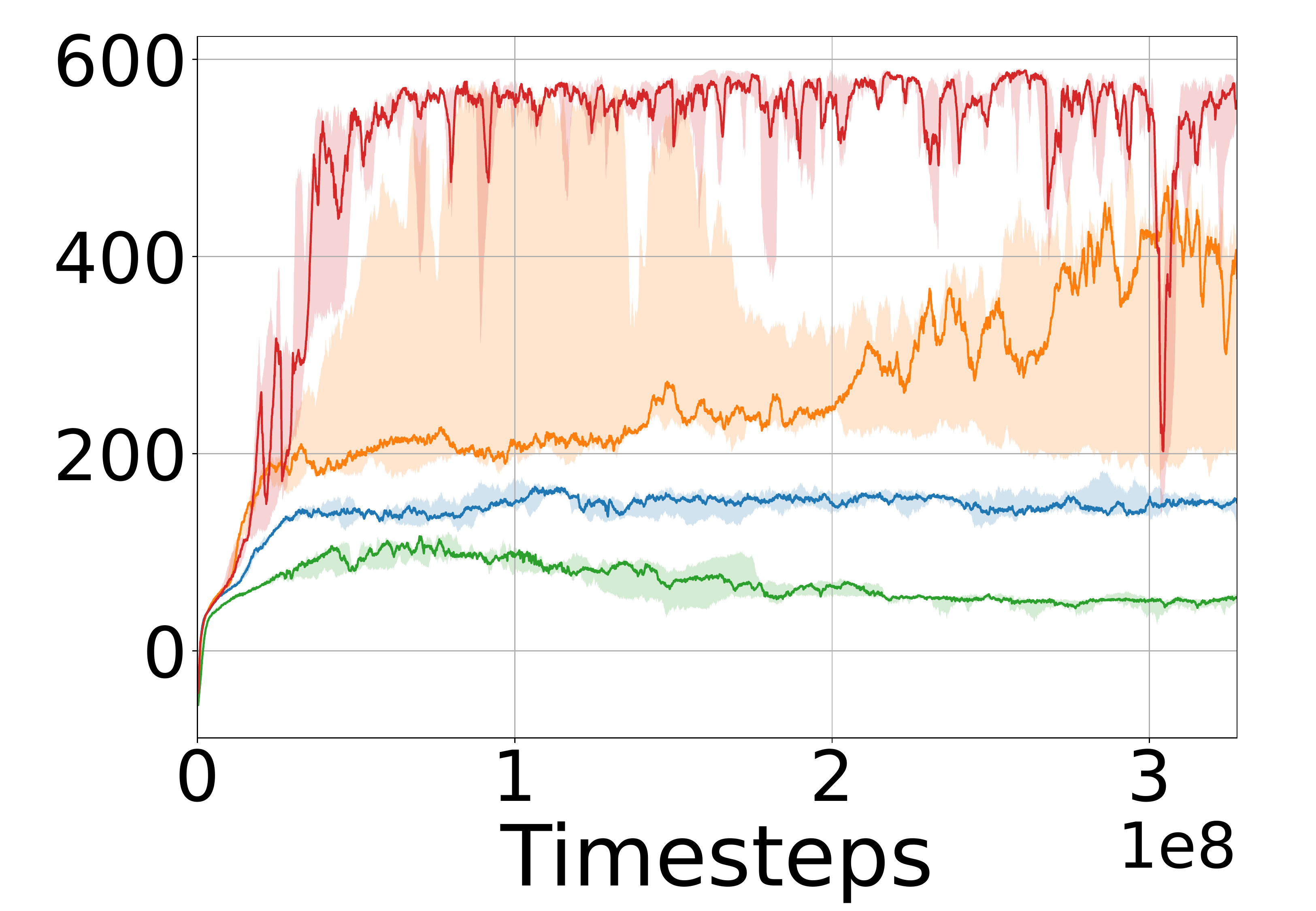}
    \caption{Ant Acrobatic}
    \label{fig:ant}
\end{subfigure}
\caption{Learning curves on robot control tasks. The x-axis is environment time steps and the y-axis is episodic returns during training. All graphs are plotted with median and 25\%-75\% percentile shading across five random seeds.} 
\label{fig:main_results}
\vspace{-5pt}
\end{figure*}

Based on Eq.~\ref{eq:equivariance}, we design the multi-agent policy and critic network structures in Fig.~\ref{fig:policy_network}, \ref{fig:critic_network}.
Agent $i$ gets a transformed observation $T_i(o)$ as the input of the policy network, the output action value consists of $a_{\text{c},i}$ and $a_{\text{s},i}$. The central joints are controlled by the mean action over all agents' output $a_{\text{c},i}$, while $a_{\text{s},i}$ will be used as the action to take for the robot part $i$. The policy network parameters are shared among agents.
The critic network gets the observations with or without actions from all agents as input. The input first goes through the shared feature learning layers in the value network.
Then the latent features are merged by a set operator \textit{mean}.
The critic value is finally calculated with the merged feature.

The proposed policy network is equivariant with respect to symmetric transformations we consider in this work, while the critic network is an invariant function.
By sharing the parameters $\Phi$ and $\Psi$ among all agents, we are able to incorporate the geometric regularization and reduce the dimension of the observation-action space.
\begin{proof}[Proof of the network equivariance/invariance]
  At the beginning we summarize the properties of the symmetry transformations in this work. They are:
    \begin{itemize}[leftmargin=*]
        \item commutative: $\mathrm{T}_j(\mathrm{T}_i(o))=\mathrm{T}_{i+j}(o)=\mathrm{T}_i(\mathrm{T}_j(o))$
        \item distributive: $\mathrm{T}_j(\mathrm{T}_i(o)+\mathrm{T}_k(o))=\mathrm{T}_j(\mathrm{T}_i(o))+\mathrm{T}_j(\mathrm{T}_k(o))$
        \item cyclic: $\mathrm{T}_i(o)=\mathrm{T}_{i+\vert\mathcal{N}\vert}(o)$
    \end{itemize}
    The equivariance of the policy for symmetric actions in Eq.~\ref{eq:equivariance} is proved as follows:
    \begin{align*}
        A_{\text{s},j}(\mathrm{T}_i(o))=\Phi_\text{s}(\mathrm{T}_{j+i}(o))=A_{\text{s},i}(\mathrm{T}_j(o))
    \end{align*}
    The equivariance for the central action is proved as follows:
    \begin{align*}
        A_{\text{c}}(\mathrm{T}_i(o))=&\frac{1}{\vert\mathcal{N}\vert}\sum_{j=0}^{\vert\mathcal{N}\vert-1}\mathrm{T}_{\vert\mathcal{N}\vert-1-j}(\Phi_\text{c}(\mathrm{T}_j(\mathrm{T}_i(o))))\\
        =&\frac{1}{\vert\mathcal{N}\vert}\sum_{j=\vert\mathcal{N}\vert-i}^{2\vert\mathcal{N}\vert-i-1}\mathrm{T}_{\vert\mathcal{N}\vert-1-j}(\Phi_\text{c}(\mathrm{T}_{i+j}(o)))\\
        =&\frac{1}{\vert\mathcal{N}\vert}\sum_{k=\vert\mathcal{N}\vert}^{2\vert\mathcal{N}\vert-1}\mathrm{T}_{\vert\mathcal{N}\vert+i-1-k}(\Phi_\text{c}(\mathrm{T}_k(o)))\\
        =&\frac{1}{\vert\mathcal{N}\vert}\sum_{k=0}^{\vert\mathcal{N}\vert-1}\mathrm{T}_i(\mathrm{T}_{\vert\mathcal{N}\vert-1-k}(\Phi_\text{c}(\mathrm{T}_k(o))))\\
        =&\mathrm{T}_i(\frac{1}{\vert\mathcal{N}\vert}\sum_{k=0}^{\vert\mathcal{N}\vert-1}\mathrm{T}_{\vert\mathcal{N}\vert-1-k}(\Phi_\text{c}(\mathrm{T}_k(o))))\\
        =&\mathrm{T}_i(A_\text{c}(o))
    \end{align*}
    The invariance of the value network is proved as follows:
    \begin{align*}
        V(\mathrm{T}_i(o))=&\Theta(\frac{1}{\vert\mathcal{N}\vert}\sum_{j=0}^{\vert\mathcal{N}\vert-1}\Psi(\mathrm{T}_j(\mathrm{T}_i(o))))\\
        =&\Theta(\frac{1}{\vert\mathcal{N}\vert}\sum_{j=|\mathcal{N}|-i}^{2\vert\mathcal{N}\vert-i-1}\Psi(\mathrm{T}_{i+j}(o)))\\
        =&\Theta(\frac{1}{\vert\mathcal{N}\vert}\sum_{k=0}^{\vert\mathcal{N}\vert-1}\Psi(\mathrm{T}_k(o)))=V(o)=V(\mathrm{T}_j(o))
    \end{align*}
    The invariance of the Q network can be proved in the same way with action $a$ concatenated to the observation $o$.
\end{proof}
 


\section{Experiments}
\label{sec:experiments_and_discussions}
We evaluate our method in different tasks to clarify the following concerns:
1) Does the multi-agent framework incorporated with robots' intrinsic symmetry improve the performance and data-efficiency on robot learning tasks?
2) Can different learning paradigms benefit from this framework?
3) Is the proposed method applicable also to real-world problems?

\subsection{Experiments with Online Reinforcement Learning}
\label{sec:online_experiments}
Previous robotic control benchmarks~\cite{tunyasuvunakool2020dmcontrol} evaluate algorithms on fundamental tasks, such as controlling agents to walk. The movements in these tasks are limited and it's relatively easy to learn a good policy. In this work, we adopt several more challenging robotic control tasks, where it is difficult for current state-of-the-art online algorithms to achieve good performance. The tasks are shown in Fig.~\ref{fig:experiments}: 
    \begin{itemize}[leftmargin=*]
        \item[] \textbf{Humanoid Tightrope}: The agent learns to control a humanoid robot to walk on a tightrope. The robot has 21 controllable motors. The tightrope is extremely narrow with a diameter of only $\SI{10}{cm}$.
        \item[] \textbf{Humanoid Dribbling}: The humanoid robot learns to dribble along routes with changing direction. Compared with the tightrope task, the observation space is augmented with features of the ball.
        \item[] \textbf{A1 Beam}: The agent controls the quadruped robot Unitree A1~\cite{unitree2018unitree} to walk on a balance beam with width of $\SI{10}{cm}$ following a predefined speed. Considering the width of A1 and the balance beam, it is much harder than walking on the ground.
        \item[] \textbf{TriFinger Lift}: TriFinger~\cite{wuthrich2021trifinger} is a 3-finger manipulator for learning dexterity. The goal is to move a cube from a random initial pose to an arbitrary 6-DoF target pose. The environment is the same as that in IssacGymEnvs~\cite{allshire2022transferring}, except that we remove the auxiliary penalty for finger movement to increase the difficulty of the task.
        \item[] \textbf{Ant Acrobatic}: The ant robot learns to do complex acrobatics (e.g., heading a pole) on a ball, which extremely challenges the ability of agents to maintain balance.
    \end{itemize}
All experiments are carried out based on the NVIDIA Isaac Gym~\cite{makoviychuk2isaac} robotics simulator.
    
\subsubsection{Baselines}
    For each task, we compare our method, named as Multi-Agent with Symmetry Augmentation (\textit{MASA}), with three baselines.
    The first baseline is Single-Agent (\textit{SA}), which treats the robot as a single agent and optimizes policy for the joint action space. This baseline can provide an intuitive comparison of our proposed framework to previous classic reinforcement learning works. The state space is kept the same as MASA's for a fair comparison.
    The second baseline is Single-Agent with Symmetry Augmentation (\textit{SASA}). It follows the SA's setup and is augmented with a symmetry loss~\cite{abdolhosseini2019learning}. Specifically, for any received observation $o$, we calculate its symmetric representation $\mathrm{T}_i(o)$. We regulate the policy function $\pi$ and the value function $V$ in PPO with extra symmetry losses by minimizing $\Vert \mathrm{T}_i(A(o))-A(\mathrm{T}_i(o))\Vert_2$ and $|V(o)-V(\mathrm{T}_i(o))|$, where $A$ and $V$ are the gathered action and critic value of the agent.
    The third baseline is Multi-Agent without Symmetry Augmentation (\textit{MA}). It uses the same architecture with parameter sharing as MASA. However, it does not involve the transformations in Fig.~\ref{fig:policy_network}~\ref{fig:critic_network}. Thus the geometric regularity of symmetry is ignored, which follows the previous research~\cite{peng2021facmac}. We concatenate a one-hot id encoding to each agent's observation as a common operation for non-homogeneous agents.
    

\subsubsection{Results}
    Figure~\ref{fig:main_results} presents the average return of all methods on different tasks during training. 
    The proposed method MASA significantly outperforms other baselines across all 5 tasks. 
    Further, the advantages over other baselines rise with the increasing difficulties of the task, which can be indicated by the increased number of joints, the extended state dimension and the enlarged state space in the task. Humanoid Tightrope and Humanoid Football control the same robot. However, in the tightrope task, the robot only needs to walk forward, while the football task involves random turns and manipulating an external object, so that other baselines can hardly learn meaningful behaviours. 
    
    By comparing the results of MASA, MA and SASA, we could observe that both of the two factors in MASA,
    multi-agent framework and symmetry structure, play an important role. Utilizing symmetry data structure alone (SASA) can gradually learn to solve a few tasks but with apparently lower data efficiency. Because the optimization space is not reduced and thus larger than that of MASA method. The multi-agent structure itself (MA) cannot guarantee meaningful results at all, which follows the criticism of naively sharing parameters among non-homogeneous agents~\cite{christianos2021scaling}.
    
    
    In the Humanoid Dribbling task, MASA initially underperforms compared to other methods. This is because the baselines prioritize self-preservation and struggle to find a policy that balances dribbling and staying alive. By focusing on avoiding falling down and kicking the ball too far away, they learn to stand still near the ball while disregarding the rewards associated with ball movement. Consequently, the baseline agents survive longer at the beginning, resulting in higher returns compared to MASA.

\subsection{Experiments with Offline Reinforcement Learning}

\begin{table*}[t]
    \vspace{5pt}
    \centering
    \begin{tabular}{lccccccc}
        \toprule
        \multirow{2}{*}{\shortstack{Datasets}} & \multirow{2}{*}{\shortstack{data}}
            & \multicolumn{3}{c}{BC} & \multicolumn{3}{c}{IQL} \\
            \cmidrule(lr){3-5} \cmidrule(lr){6-8}
           & & SA & SASA-data & MASA & SA & SASA-data & MASA \\
         \midrule
         Sim-Expert & 0.87 & $0.64\pm0.00$ & $0.29\pm0.13$ & $\bm{0.71\pm0.05}$ & $0.47\pm0.06$ & $0.37\pm0.13$ & $\bm{0.84\pm0.02}$ \\
         Sim-Half-Expert & 0.88 & $0.64\pm0.02$ & $0.37\pm0.08$ & $\bm{0.75\pm0.05}$ & $0.04\pm0.01$ & $0.10\pm0.05$ & $\bm{0.78\pm0.06}$ \\
         Sim-Weak\&Expert & 0.5 & $0.16\pm0.04$ & $0.10\pm0.05$ & $\bm{0.34\pm0.07}$ & $0.24\pm0.05$ & $0.20\pm0.07$ & $\bm{0.55\pm0.04}$ \\
         Sim-Mixed & 0.68 & $0.01\pm0.01$ & $0.01\pm0.01$ & $\bm{0.23\pm0.10}$ & $0.00\pm0.00$ & $0.03\pm0.02$ & $\bm{0.48\pm0.09}$ \\
         \cmidrule(lr){2-8}
         Real-Expert & 0.66 & $0.27\pm0.09$ & $0.13\pm0.15$ & $\bm{0.52\pm0.18}$ & $0.29\pm0.09$ & $0.16\pm0.18$ & $\bm{0.61\pm0.18}$ \\
         Real-Half-Expert & 0.68 & $0.15\pm0.01$ & $0.13\pm0.14$ & $\bm{0.41\pm0.25}$ & $0.12\pm0.06$ & $0.11\pm0.13$ & $\bm{0.61\pm0.19}$ \\
         Real-Weak\&Expert & 0.40 & $0.02\pm0.04$ & $0.05\pm0.09$ & $\bm{0.26\pm0.20}$ & $0.11\pm0.13$ & $0.17\pm0.14$ & $\bm{0.30\pm0.24}$ \\
         Real-Mixed & 0.42 & $0.00\pm0.01$ & $0.04\pm0.09$ & $\bm{0.06\pm0.08}$ & $0.03\pm0.02$ & $0.03\pm0.08$ & $\bm{0.22\pm0.15}$ \\
        \bottomrule
    \end{tabular}
    \caption{Success rate on the TriFinger-Lift datasets. Average and standard deviation over five training seeds. `data' denotes the mean over the dataset.}
    \label{tab:offline}
    \vspace{-5pt}
\end{table*}

\label{sec:offline_experiments}
    Although reinforcement learning enables agent to learn from interactions with the environment, collecting samples during training could be inefficient or unsafe~\cite{arnold2020emp}.
    Offline learning solves this problem by learning from fixed dataset.
    To demonstrate our method's generalizability, we also applied it to behavior cloning and implicit Q-learning, and evaluate it on both a simulator and real robots.

\subsubsection{System Setup}
    The work~\cite{gurtler2023benchmarking} publishes large diverse datasets collected on a Pybullet simulator and a real robot cluster for benchmarking offline reinforcement learning algorithms.
    The simulator is provided and the robot cluster can be accessed remotely for evaluating learned policies.
    We choose the more complex task \textbf{Lift} in the dataset to show the potential of our method.
    The task is basically the same as \textit{TriFinger Lift} described in Sec.~\ref{sec:online_experiments}.

\subsubsection{Baselines}
    We evaluate MASA against two baselines: SA and SASA-data.
    The first baseline is the original offline algorithm treating the robot as a single agent and learn policies for the joint action space.
    The SASA-data baseline is different from SASA in Sec.~\ref{sec:online_experiments}, which add auxiliary loss functions to the policy and critic learning. It first augments the offline dataset with symmetric transitions $\big(\mathrm{T}_i(o),\mathrm{T}_i(a)\big)$. Then an agent is trained with SA method on the larger dataset.
    We do not benchmark against the naive multi-agent method MA due to its poor performance shown in Sec.~\ref{sec:online_experiments}.

\subsubsection{Evaluation}
    The evaluation results for the TriFinger Lift task on both simulator and real-robots are summarized in Table~\ref{tab:offline}.
    The values for SA come from the benchmark paper~\cite{gurtler2023benchmarking}. To keep fairness, we use the default hyperparameters for training all policies.
    The evaluation process are also kept unchanged.
    The policies trained on simulation datasets are evaluated in the corresponding simulator. $100$ episodes are carried out for each seed and algorithm.
    Those trained on real-robot datasets are evaluated with the real-robot cluster. We run $6$ episodes for each seed and algorithm.
    The final success rate values are averaged across five training seeds, which are also the same as those in the benchmark paper.
    
    Our method outperforms the baselines over all datasets with a significant margin without any hyperparameter fine-tuning.
    For more than half of the datasets, the MASA agents achieve success rates close to or even better than that of the data collection policy.
    We observe two interesting results:
    I) The relative performance gain of MASA on most real-robot datasets is higher than that on simulation datasets.
    This can be explained as a result of the enhanced domain randomization with MASA. The robot in simulator is perfect and requires no domain randomization. The real robots, however, have different characteristics for different robots and fingers. Assuming $k$ different robots are in the dataset, a MASA agent learns the experience from $3\times k$ fingers due to the shared policy.
    II) The SASA-data agents underperform the SA agents.
    We think the performance drop is caused by ambiguous regression targets introduced by data augmentation. Given two samples $(o,a)$ and $(o',a')$ in the original dataset, where $o'\approx \mathrm{T}_1(o),a'\neq \mathrm{T}_1(a)$, the resulting augmented data elements would be $(o,a),\big(\mathrm{T}_1(o),\mathrm{T}_1(a)\big),\big(\mathrm{T}_2(o),\mathrm{T}_2(a)\big)$ and $\big(o,\mathrm{T}_2(a')\big),\big(\mathrm{T}_1(o),a'\big),\big(\mathrm{T}_2(o),\mathrm{T}_1(a')\big)$, which means different actions for similar observations. Since the data collection policy is not trained with a symmetric policy, such multi-modality in control sequences can be introduced by symmetric transformation. As a result, BC agents tend to learn out-of-distribution actions with such ambiguous regression targets. This problem is less severe for offline reinforcement learning algorithms like IQL, because they often are equipped with regularizers to avoid out-of-distribution actions.

\section{Conclusion and Limitations}
\label{sec:conclusion}
This paper introduced a novel approach that incorporates the robot's intrinsic symmetry into an agent's policy and critic networks. Through our unique network architecture, MASA, we demonstrated remarkable success in learning robot control tasks, integrating our method with PPO, BC, and IQL from different learning paradigms. Despite the imperfections and variances in real-world robots due to mounting and manufacturing tolerances, our approach still enhances learning algorithms, even leveraging these imperfections to its advantage.
Although the implementation of our method necessitates specific domain knowledge, such as understanding the robot structure and the transformation operations, it provides a significant contribution to robot learning in complex tasks. Furthermore, MASA serves as a valuable blueprint for developing robots with increased degrees of freedom, all the while keeping the complexity of observation-action space manageable. Future work offers promising avenues, including the exploration of additional symmetric structures and the automation of identifying robots' intrinsic symmetries. The question how varying degrees of symmetry imperfection affect our method's performance is also an interesting aspect for future work.

\section*{Acknowledgement}
We extend our gratitude to the Max Planck Institute for Intelligent Systems in Tübingen, Germany, for providing the TriFinger robots, encompassing software, hardware, and datasets.
Special thanks to Nico Gürtler, whose support and invaluable discussions significantly contributed to our experiments.




%
%

\bibliographystyle{IEEEtran}
\bibliography{masa}

\appendix
    

\subsection{Extra Experimental Setups}
\label{sec:extra_experimental_results}

\subsubsection{Hyperparameters}
\label{sec:hyperparameters}
    Each baseline is run with 5 random seeds. All experiments are carried out on GPU card NVIDIA rtxA6000 and rtx3080 GPU. 
    The hyperparameters of all baselines are consistent for a fair comparison. The detailed values can be accessed in Table~\ref{tab:hyperparameters}.

    \begin{table*}
    \caption{Hyperparameters of all experiments. }
    \begin{center}
        \begin{small}
            \begin{sc}
              \scalebox{0.9}{
                \begin{tabular}{lcccccc}
                \toprule
                \textbf{Hyperparameters} & \textbf{Humanoid Tightrope} & \textbf{Humanoid Football} & 
                \textbf{Trifinger Move} & 
                \textbf{A1 Beam} & 
                \textbf{Ant Acrobatic} & 
                \\
                \midrule
                batch size & 4096$\times$32 & 4096$\times$32 & 16384$\times$16 & 4096$\times$24 & 4096$\times$16 \\
                        
                mixed precision & True & True & False & True & True \\
                normalize input & True & True & True & True & True \\
                normalize value & True & True & True & True & True \\
                value bootstrap & True & True & True & True & True \\
                num actors & 4096 & 4096 & 16384 & 4096 & 4096 \\
                normalize advantage & True & True & True & True & True \\
                gamma  & 0.99 & 0.99 & 0.99 & 0.99 & 0.99 \\
                tau  & 0.95 & 0.95 & 0.95 & 0.95 & 0.95 \\
                e-clip  & 0.2 & 0.2 & 0.2 & 0.2 & 0.2 \\
                entropy coefficient & 0.0  & 0.0  & 0.0  & 0.0  & 0.0 \\
                learning rate & 5.e-4  & 5.e-4  & 3.e-4  & 3.e-4  & 3.e-4 \\
                kl threshold & 0.0008 & 0.0008 & 0.0008 & 0.0008 & 0.0008\\
                truncated grad norm & 1.0  & 1.0  & 1.0  & 1.0  & 1.0 \\
                horizon length & 32  & 32  & 16  & 24  & 16 \\
                minibatch size & 32768  & 32768  & 16384  & 32768  & 32768 \\
                mini epochs & 5  & 5  & 4  & 5 & 4 \\
                critic coefficient  & 4.0  & 4.0  & 4.0  & 2.0 & 2.0 \\
                max epoch  & 10k  & 10k  & 10k  & 10k & 5k \\
                policy network & [400,200,100] & [400,200,100] & [256,256,128,128] & [256, 128, 64] & [256, 128, 64] \\
                critic network & [400,200,100] & [400,200,100] & [256,256,128,128] & [256, 128, 64] & [256, 128, 64] \\
                activation function & ELU & ELU & ELU & ELU & ELU \\
                \bottomrule 
                \end{tabular}
                }
            \end{sc}
        \end{small}
    \end{center}
    \label{tab:hyperparameters}
\end{table*}

\subsubsection{Tasks Details}
\label{sec:task_details}
        
    \paragraph{Humanoid Tightrope}
    In this task, the agent learns to control a humanoid robot to walk on a tightrope. The humanoid robot has 21 controllable motors. The tightrope is extremely narrow with a diameter of only $\SI{10}{cm}$, which challenges the efficiency of learning algorithms. The agent is rewarded with a forward speed on the tightrope and a proper posture.
    At each non-terminating step, the reward $r=w_v\times r_v+w_\text{alive}\times r_\text{alive}+w_\text{up}\times r_\text{up}+w_\text{heading}\times r_\text{heading}+w_\text{action}\times r_\text{action}+ w_\text{energy}\times r_\text{energy}+w_{\text{lateral}}\times r_\text{lateral}$, where
    \begin{itemize}
        \item $r_v$ is the robot's forward velocity, $w_v=1.0$;
        \item $r_\text{alive}=1$, $w_\text{alive}=2.0$;
        \item $r_\text{up}=1$ if $e_{\text{up},z}>0.93$, where $e_\text{up}$ is the basis vector of torso's $z$ axis in the global coordinate system, otherwise the value is $0$, $w_\text{up}=0.1$;
        \item $r_\text{heading}=e_{\text{forward},x}$, where $e_\text{forward}$ is the basis vector of torso's $x$ axis in global coordinate system, $w_\text{forward}=0.1$;
        \item $r_\text{action}=\Vert a\Vert_2^2$, where $a$ is joints action, $w_\text{action}=-0.01$
        \item $r_\text{energy}$ is the joints power consumption, $w_\text{energy}=-0.05$
        \item $r_\text{lateral}=v_{\text{torso},y}$ is the penalty for lateral velocity, $w_{\text{lateral}}=-1.0$
    \end{itemize}
    The reward is $-1$ for termination step. The action is the force applied to all joints.
        
    \paragraph{Humanoid Dribbling}
    In this task, the robot learns to dribble along routes with random turns. The observation space is augmented with features of the ball compared with the tightrope task. For observation calculation, the global coordinate system changes with the new target route at the turning position.
    At each non-terminating step, the reward $r=w_v\times r_v+w_\text{alive}\times r_\text{alive}+w_\text{dist}\times r_\text{dist}+w_\text{heading}\times r_\text{heading}+w_\text{action}\times r_\text{action}+ w_\text{energy}\times r_\text{energy}+w_{\text{lateral}}\times r_\text{lateral}$, where
    \begin{itemize}
        \item $r_v$ is the ball's forward velocity, $w_v=2.0$;
        \item $r_\text{alive}=1$, $w_\text{alive}=0.2$;
        \item $r_\text{dist}=e^{-d}$ where $d$ is the 2d distance from torso to the ball, $w_\text{dist}=0.2$;
        \item $r_\text{heading}=e_{\text{forward},x}$, where $e_\text{forward}$ is the basis vector of torso's $x$ axis in the global system, $w_\text{forward}=1.0$;
        \item $r_\text{action},r_\text{energy}$ are the same with Humanoid Tightrope
        \item $r_\text{lateral}=v_{\text{ball},y}$ is the penalty for the ball's lateral velocity, $w_{\text{lateral}}=-0.5$
    \end{itemize}
    The reward is $-1$ for termination step. The action is the force applied to all joints.
    
    \paragraph{A1 Beam}
    In this task, the agent controls the quadruped robot Unitree A1~\cite{unitree2018unitree} to walk on a balance beam with width of $\SI{10}{cm}$ following a predefined speed. Considering the width of A1 and the balance beam, it is much harder than walking on the ground. There are overall 12 motors for Unitree A1, 3 for each leg.
    At each non-terminating step, the reward $r=w_v\times r_v+w_\text{alive}\times r_\text{alive}+w_\text{heading}\times r_\text{heading}+w_\text{action}\times r_\text{action}+ w_{\text{lateral}}\times r_\text{lateral}$, where
    \begin{itemize}
        \item $r_v=e^{-|v_{\text{torso},x}-v_\text{target}|}$ is speed tracking reward, $w_v=1.0$;
        \item $r_\text{alive}=1$, $w_\text{alive}=1.0$;
        \item $r_\text{heading}=e_{\text{forward},x}$, where $e_\text{forward}$ is the basis vector of torso's $x$ axis in global coordinate system, $w_\text{forward}=1.0$;
        \item $r_\text{action}=\Vert a\Vert_2^2$, where $a$ is the joints action, $w_\text{action}=-0.5$
        \item $r_\text{lateral}=v_{\text{torso},y}$ is penalty for lateral velocity, $w_{\text{lateral}}=-1.0$
    \end{itemize}
    The reward is $-1$ for termination step. The robot has a low-level joint controller. The action is the target angular position of all joints.

    \paragraph{Trifinger Lift}
    Trifinger~\cite{wuthrich2021trifinger} is a 3-finger manipulator for learning dexterity. The goal of the task is to move a cube from a random initial pose to an arbitrary 6-DoF target position and orientation. The environment is the same as that of~\cite{allshire2022transferring}, except that we remove the auxiliary penalty for finger movement, which increases the difficulty of the task. The robot has a low-level joint controller. The action is the target angular position of all joints.
        
    \paragraph{Ant Acrobatic} 
    In this task, an ant learns to do complex acrobatics (e.g. heading a pole) on a ball, which extremely challenges the ability of agents to maintain balance. The action space is 8 dimensions. 
    At each non-terminating step, the reward $r=w_\text{alive}\times r_\text{alive}+w_\text{action}\times r_\text{action}+w_\text{energy}\times r_\text{energy}$, where
    \begin{itemize}
        \item $r_\text{alive}=1$, $w_\text{alive}=0.5$;
        \item $r_\text{action}=\Vert a\Vert_2^2$, where $a$ is joints action, $w_\text{action}=-0.005$
        \item $r_\text{energy}$ is joints power consumption, $w_\text{energy}=-0.05$
    \end{itemize}
    The reward is $-1$ for termination step. The action is the force applied to all joints.

    We conclude the observation space for each task in Table~\ref{tab:task_info} for easier reading.

    \begin{table*}
    \caption{Tasks Information}
    \begin{center}
        \begin{small}
            \begin{sc}
              \scalebox{0.9}{
                \begin{tabular}{c|cccccc}
                \toprule
                \multicolumn{2}{l}{} & \textbf{Humanoid Tightrope} & \textbf{Humanoid Football} & 
                \textbf{Trifinger Move} & 
                \textbf{A1 Beam} & 
                \textbf{Ant Acrobatic}
                \\
                \midrule
                \multicolumn{2}{l}{observation dimension} & 74 & 80 & 41 & 47 & 57 \\
                \midrule
                \multirow{42}{*}{$o_{\text{c}}$} & \multirow{12}{*}{torso} & $y_\text{torso}$ & $y_\text{torso}$ & & $y_\text{torso}$ & $x_\text{torso}$ \\
                & & $z_\text{torso}$ & $z_\text{torso}$ & & $z_\text{torso}$ & $y_\text{torso}$ \\
                & & $v_{\text{torso},x}$ & $v_{\text{torso},x}$ & & $v_{\text{torso},x}$ & $z_\text{torso}$ \\
                & & $v_{\text{torso},y}$ & $v_{\text{torso},y}$ & & $v_{\text{torso},y}$ & $v_{\text{torso},x}$ \\
                & & $v_{\text{torso},z}$ & $v_{\text{torso},z}$ & & $v_{\text{torso},z}$ & $v_{\text{torso},y}$ \\
                & & $\omega_{\text{torso},x}$ & $\omega_{\text{torso},x}$ & & $\omega_{\text{torso},x}$ & $v_{\text{torso},z}$ \\
                & & $\omega_{\text{torso},y}$ & $\omega_{\text{torso},y}$ & & $\omega_{\text{torso},y}$ & $\omega_{\text{torso},x}$ \\
                & & $\omega_{\text{torso},z}$ & $\omega_{\text{torso},z}$ & & $\omega_{\text{torso},z}$ & $\omega_{\text{torso},y}$ \\
                & & $\alpha_{\text{torso}}$ & $\alpha_{\text{torso}}$ & & $\alpha_{\text{torso}}$ & $\omega_{\text{torso},z}$ \\
                & & $\beta_{\text{torso}}$ & $\beta_{\text{torso}}$ & & $\beta_{\text{torso}}$ & $\alpha_{\text{torso}}$ \\
                & & $\gamma_{\text{torso}}$ & $\gamma_{\text{torso}}$ & & $\gamma_{\text{torso}}$ & $\beta_{\text{torso}}$ \\
                & & & & & & $\gamma_{\text{torso}}$ \\
                \cline{2-7}
                & \multirow{9}{*}{torso joints} & $\theta_{\text{lower waist},x}$ & $\theta_{\text{lower waist},x}$ & & & \\
                & & $\theta_{\text{lower waist},y}$ & $\theta_{\text{lower waist},y}$ & & & \\
                & & $\theta_{\text{pelvis},x}$ & $\theta_{\text{pelvis},x}$ & & & \\
                & & $\omega_{\text{lower waist},x}$ & $\omega_{\text{lower waist},x}$ & & & \\
                & & $\omega_{\text{lower waist},y}$ & $\omega_{\text{lower waist},y}$ & & & \\
                & & $\omega_{\text{pelvis},x}$ & $\omega_{\text{pelvis},x}$ & & & \\
                & & $a_{\text{lower waist},x}$ & $a_{\text{lower waist},x}$ & & & \\
                & & $a_{\text{lower waist},y}$ & $a_{\text{lower waist},y}$ & & & \\
                & & $a_{\text{pelvis},x}$ & $a_{\text{pelvis},x}$ & & & \\
                \cline{2-7}
                & \multirow{21}{*}{external objects} & & $x_\text{ball}$ & $x_\text{cube}$ & & $x_\text{pole}$ \\
                & & & $y_\text{ball}$ & $y_\text{cube}$ & & $y_\text{pole}$ \\
                & & & $z_\text{ball}$ & $z_\text{cube}$ & & $z_\text{pole}$ \\
                & & & $v_{\text{ball},x}$ & $H_{\text{cube},x}$ & & $v_{\text{pole},x}$ \\
                & & & $v_{\text{ball},y}$ & $H_{\text{cube},y}$ & & $v_{\text{pole},y}$ \\
                & & & $v_{\text{ball},z}$ & $H_{\text{cube},z}$ & & $v_{\text{pole},z}$ \\
                & & & & $H_{\text{cube},w}$ & & $\omega_{\text{pole},x}$ \\
                & & & & $x_\text{cube target}$ & & $\omega_{\text{pole},y}$ \\
                & & & & $y_\text{cube target}$ & & $\omega_{\text{pole},z}$ \\
                & & & & $z_\text{cube target}$ & & $\text{up}_{\text{pole},x}$ \\
                & & & & $H_{\text{cube target},x}$ & & $\text{up}_{\text{pole},y}$ \\
                & & & & $H_{\text{cube target},y}$ & & $\text{up}_{\text{pole},z}$ \\
                & & & & $H_{\text{cube target},z}$ & & $x_\text{ball}$ \\
                & & & & $H_{\text{cube target},w}$ & & $y_\text{ball}$ \\
                & & & & & & $z_\text{ball}$ \\
                & & & & & & $v_{\text{ball},x}$ \\
                & & & & & & $v_{\text{ball},y}$ \\
                & & & & & & $v_{\text{ball},z}$ \\
                & & & & & & $\omega_{\text{ball},x}$ \\
                & & & & & & $\omega_{\text{ball},y}$ \\
                & & & & & & $\omega_{\text{ball},z}$ \\
                \cline{1-7}
                \multirow{27}{*}{$o_{\text{s},i}$} & \multirow{27}{*}{limb joints} & $\theta_{\text{upper arm},x}$ & $\theta_{\text{upper arm},x}$ & $\theta_{\text{finger upper}}$ & $\theta_{\text{front hip}}$ & \\
                & & $\theta_{\text{upper arm},z}$ & $\theta_{\text{upper arm},z}$ & $\theta_{\text{finger middle}}$ & $\theta_{\text{front thigh}}$ & \\
                & & $\theta_{\text{lower arm},x}$ & $\theta_{\text{lower arm},x}$ & $\theta_{\text{finger lower}}$ & $\theta_{\text{front calf}}$ & \\
                & & $\theta_{\text{thigh},x}$ & $\theta_{\text{thigh},x}$ & $\omega_\text{finger upper}$ & $\theta_{\text{rear hip}}$ & \\
                & & $\theta_{\text{thigh},y}$ & $\theta_{\text{thigh},y}$ & $\omega_\text{finger middle}$ & $\theta_{\text{rear thigh}}$ & \\
                & & $\theta_{\text{thigh},z}$ & $\theta_{\text{thigh},z}$ & $\omega_\text{finger lower}$ & $\theta_{\text{rear calf}}$ & \\
                & & $\theta_{\text{knee},x}$ & $\theta_{\text{knee},x}$ & $a_\text{finger upper}$ & $\omega_{\text{front hip}}$ & \\
                & & $\theta_{\text{foot},x}$ & $\theta_{\text{foot},x}$ & $a_\text{finger middle}$ & $\omega_{\text{front thigh}}$ & \\
                & & $\theta_{\text{foot},y}$ & $\theta_{\text{foot},y}$ & $a_\text{finger lower}$ & $\omega_{\text{front calf}}$ & \\
                & & $\omega_{\text{upper arm},x}$ & $\omega_{\text{upper arm},x}$ & & $\omega_{\text{rear hip}}$ & \\
                & & $\omega_{\text{upper arm},z}$ & $\omega_{\text{upper arm},z}$ & & $\omega_{\text{rear thigh}}$ & \\
                & & $\omega_{\text{lower arm},x}$ & $\omega_{\text{lower arm},x}$ & & $\omega_{\text{rear calf}}$ & \\
                & & $\omega_{\text{thigh},x}$ & $\omega_{\text{thigh},x}$ & & $a_{\text{front hip}}$ & \\
                & & $\omega_{\text{thigh},y}$ & $\omega_{\text{thigh},y}$ & & $a_{\text{front thigh}}$ & \\
                & & $\omega_{\text{thigh},z}$ & $\omega_{\text{thigh},z}$ & & $a_{\text{front calf}}$ & \\
                & & $\omega_{\text{knee},x}$ & $\omega_{\text{knee},x}$ & & $a_{\text{rear hip}}$ & \\
                & & $\omega_{\text{foot},x}$ & $\omega_{\text{foot},x}$ & & $a_{\text{rear thigh}}$ & \\
                & & $\omega_{\text{foot},y}$ & $\omega_{\text{foot},y}$ & & $a_{\text{rear calf}}$ & \\
                & & $a_{\text{upper arm},x}$ & $a_{\text{upper arm},x}$ & & & \\
                & & $a_{\text{upper arm},z}$ & $a_{\text{upper arm},z}$ & & & \\
                & & $a_{\text{lower arm},x}$ & $a_{\text{lower arm},x}$ & & & \\
                & & $a_{\text{thigh},x}$ & $a_{\text{thigh},x}$ & & & \\
                & & $a_{\text{thigh},y}$ & $a_{\text{thigh},y}$ & & & \\
                & & $a_{\text{thigh},z}$ & $a_{\text{thigh},z}$ & & & \\
                & & $a_{\text{knee},x}$ & $a_{\text{knee},x}$ & & & \\
                & & $a_{\text{foot},x}$ & $a_{\text{foot},x}$ & & & \\
                & & $a_{\text{foot},y}$ & $a_{\text{foot},y}$ & & & \\
                \midrule
                \multicolumn{2}{l}{$|\mathcal{N}|$} & 2 & 2 & 3 & 2 & 4 \\
                \midrule
                \multicolumn{2}{l}{action dimension}  & 21 & 21 & 9 & 12 & 8 \\
                \bottomrule 
                \end{tabular}
                }
            \end{sc}
        \end{small}
    \end{center}
    \label{tab:task_info}
\end{table*}


\end{document}